\documentclass[10pt,twocolumn,letterpaper]{article}

\usepackage[pagenumbers]{iccv} % To force page numbers, e.g. for an arXiv version

\definecolor{iccvblue}{rgb}{0.21,0.49,0.74}
\usepackage[pagebackref,breaklinks,colorlinks,allcolors=iccvblue]{hyperref}
\usepackage{tikz}
\usetikzlibrary{bayesnet, positioning}
\usepackage{subcaption}
\usepackage{xcolor}
\usepackage{multirow}
\usepackage{xspace}
\usepackage[accsupp]{axessibility}
\usepackage{tabularx}

\newcommand{\para}[1]{\noindent\textbf{#1}}

\newcommand{\ourmethod}{MOSAIC\xspace}

\def\paperID{10251} % *** Enter the Paper ID here
\def\confName{ICCV}
\def\confYear{2025}

\title{\ourmethod: Generating Consistent, Privacy-Preserving Scenes \\ from Multiple Depth Views in Multi-Room Environments}
\author{Zhixuan Liu$^{1}$ \and Haokun Zhu$^{1}$ \and Rui Chen$^{1}$ \and  Jonathan Francis$^{1,2}$ \and Soonmin Hwang$^{3}$ \quad Ji Zhang$^{1}$ \quad Jean Oh$^{1}$ \\
$^1$Carnegie Mellon University \quad
$^2$Bosch Center for AI \quad
$^3$Hanyang University
}

\begin{document}
\twocolumn[{%
\renewcommand\twocolumn[1][]{#1}
\maketitle
\begin{center}
    \vspace{-0.5cm}
    \centering \small
    \captionsetup{type=figure}
    \includegraphics[width=0.9\textwidth]{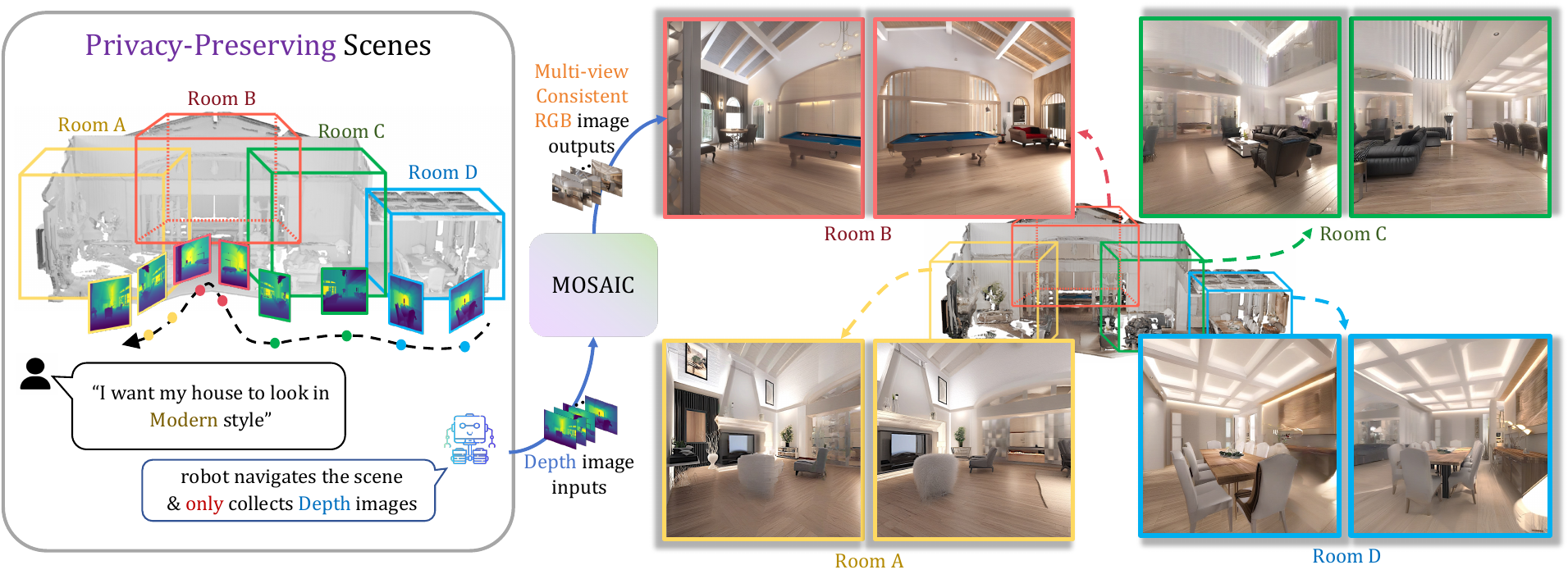}%
    \vspace{-10.2pt}
    \captionof{figure}{{For privacy-preserving scenarios where RGB collection is restricted, MOSAIC generates consistent RGB images from depth data captured along robot paths, guided by text prompts. These outputs further enable 3D reconstruction of multi-room environments.}}%
    \label{fig:teaser}%
\end{center}%
}]
\begin{abstract}
% We introduce a zero-shot framework that enables autonomous robots to generate privacy-preserving digital twins of indoor environments without recording any RGB data.
% In contrast to conventional scene reconstruction pipelines, which rely on color images and thus raise privacy concerns
We introduce a diffusion-based approach for generating privacy-preserving digital twins of multi-room indoor environments from depth images only.
% our approach collects only depth information and converts key-frame sets into photorealistic, multi-view-consistent RGB renderings via a depth-conditioned diffusion model. 
Central to our approach is a novel \textbf{M}ulti-view \textbf{O}verlapped \textbf{S}cene \textbf{A}lignment with \textbf{I}mplicit \textbf{C}onsistency (\ourmethod) model that explicitly considers cross-view dependencies within the same scene in the probabilistic sense.
\ourmethod operates through a multi-channel inference-time optimization that avoids error accumulation common in sequential or single-room constraints in panorama-based approaches.
\ourmethod scales to complex scenes with zero extra training and provably reduces the variance during denoising process when more overlapping views are added, leading to improved generation quality.
Experiments show that \ourmethod outperforms state-of-the-art baselines on image fidelity metrics in reconstructing complex multi-room environments. Resources and code are at \url{https://mosaic-cmubig.github.io}.
% with irregular structures.
% We also introduce a text-guided appearance control mechanism that enables on-the-fly style customization, making \ourmethod a flexible end-to-end solution for hospitals, elder care facilities, factories, and other sensitive settings where RGB data collection is restricted. it Crucially, NavGen employs while—supporting 
\end{abstract}    
\vspace{-0.5cm}
\section{Introduction}
\label{sec:intro}
Autonomous scene reconstruction is a crucial capability in robotics and computer vision~\cite{mildenhall2020nerfrepresentingscenesneural, murez2020atlas, kerbl20233dgaussiansplattingrealtime, amatare2024dtradardigitaltwinassisted}, with applications spanning virtual reality, architectural design, and AI-driven simulation. However, existing multi-view 3D reconstruction methods rely heavily on capturing RGB images, which poses significant privacy risks in sensitive environments such as hospitals, factories, elder care facilities, and private residences. Capturing RGB data in these settings can inadvertently expose personal information~\cite{ravi2021reviewvisualprivacypreservation, zhao2022privacypreservingreflectionrenderingaugmented}, limiting the practicality of generative AI solutions for real-world deployments where visual privacy must be preserved.

A promising solution for privacy-sensitive environments involves deploying mobile robots that collect only geometric data, such as depth images: this strategy preserves the scene's structural layout while preventing the capture of sensitive texture information. In order to create digital replicas of these scenes, we need algorithms that are capable of generating high-quality, multiview-consistent sequential images that align with ground truth geometry, while maintaining visual coherence across viewpoints.

Various approaches for scene-level multi-view image generation have been proposed, with autoregressive generation techniques \cite{hollein2023text2room, fridman2023scenescape, yu2024wonderjourneygoing, yu2024wonderworldinteractive3dscene, chung2023luciddreamer} representing the current mainstream trend. These approaches leverage powerful inpainting models to iteratively render unseen parts of the scene from sequential camera perspectives. However, these approaches exhibit style drift, where images generated for initial viewpoints often differ stylistically from those produced when revisiting the same location from different angles, creating visual inconsistencies. Moreover, warp-and-inpainting methods frequently encounter depth misalignment issues, %even in situations where ground truth depth alignment is a requirement, 
leading to compounding errors over sequential generations~\cite{song2023roomdreamertextdriven3dindoor, controlnetinpaint2023}. %Although some image-generation approaches attempt to address these misalignment issues by distilling a depth-conditioned inpainting model~\cite{yang2025scenecraft}, they still struggle with image quality and generalization challenges, remain limited to simple single-room environments, and thus lack robustness for real-world applications with complex multi-room layouts. 
Unlike prior methods that perform diffusion synchronization tasks~\cite{bartal2023multidiffusionfusingdiffusionpaths,lee2023syncdiffusioncoherentmontagesynchronized,geng2024visualanagramsgeneratingmultiview,liu2023textguidedtexturingsynchronizedmultiview,kim2024synctweediesgeneralgenerativeframework} in constrained settings (limited viewpoint variations), %%: projections, orthogonal transforms, or UV mappings at the object level with limited viewpoint variations. 
our task requires coherence between viewpoints based on arbitrary trajectory features, extensive perspective changes, and spatial discontinuities. This fundamental distinction makes existing methods ineffective for complex multi-room environments with widely-varying camera positions. %%Simply averaging across diffusion denoising levels fails to preserve consistency in these challenging scenarios. 

To overcome these limitations, we present \textbf{MOSAIC}
(\textbf{M}ulti‑view \textbf{O}verlapped \textbf{S}cene \textbf{A}lignment with \textbf{I}mplicit \textbf{C}onsistency),
a training‑free diffusion framework that converts privacy‑preserving depth sequences into photorealistic,
geometry‑aligned RGB views across large, cluttered, multi‑room environments.
Previous work~\cite{tang2023mvdiffusion} enforces consistency by
fine‑tuning a modified architecture on trajectory‑specific data, limiting scalability and
generalization; MOSAIC instead performs a lightweight multi‑view inference‑time optimization that scales to unseen environments
and enforces cross‑view agreement at every denoising step, while preserving the intrinsic knowledge of a large pre‑trained model~\cite{podell2023sdxlimprovinglatentdiffusion}.
A depth‑weighted projection loss highlights the most reliable viewpoints,
and a final pixel‑space refinement converts latent‑space coherence into precisely aligned images.
As more overlapping views are added, MOSAIC reduces variance and tightens image--depth alignment, unlike autoregressive pipelines that compound errors, thus delivering stable reconstructions over arbitrarily long trajectories.
Our experiments demonstrate superior performance over existing methods across multiple metrics, establishing a new state-of-the-art for scene level multi-view image generation with geometric priors.

In summary, (1) we introduce \ourmethod, a training-free diffusion pipeline that turns arbitrary depth sequences into photorealistic geometry-aligned RGB views, (2) we devise an inference-time sampler, with depth-weighted projection loss and late pixel refinement, that enforces strict cross-view coherence, (3) we theoretically prove that adding overlapping views monotonically reduces denoising variance, curbing error accumulation, and (4) we demonstrate state-of-the-art results in fidelity, prompt-following, and geometry alignment in cluttered, multi-room environments.

\section{Related Work}
\label{sec:relatedwork}
\para{Diffusion‑guided scene‑level multi‑view generation.}
Recent advances in 2D diffusion-based generative models~\cite{ddim, ddpm, rombach2022high, control_net, podell2023sdxlimprovinglatentdiffusion, esser2024scalingrectifiedflowtransformers} have catalyzed significant progress in scene-level multiview image generation, yet most pipelines rely on iterative warping and inpainting for view completion~\cite{hollein2023text2room,fridman2023scenescape,yu2024wonderjourneygoing,yu2024wonderworldinteractive3dscene,chung2023luciddreamer},
where style drift accumulates across iterations and produces artifacts.
A panorama‑first strategy mitigates drift by re‑projecting a single generated panorama~\cite{song2023roomdreamertextdriven3dindoor,li2024scenedreamer360textdriven3dconsistentscene,schult2023controlroom3droomgenerationusing,wang2024roomtextexturingcompositionalindoor},
but cannot cope with multi‑room layouts or long, unconstrained trajectories.
MVDiffusion~\cite{tang2023mvdiffusion} requires training an additional module for cross-view consistency, which limits the generalization to unseen environments and prompt following fidelity.
SceneTex~\cite{chen2023scenetexhighqualitytexturesynthesis} assumes a pre‑meshed scene and performs score‑distillation texture optimization, requiring $\sim$20 hours per scene and often seeking to a single appearance mode; our training‑free DDIM process finishes in few minutes while preserving generation diversity under single text prompt.
Recent video‑diffusion models provide temporal coherence~\cite{svd,ho2022video,yang2024cogvideoxtexttovideodiffusionmodels,zheng2024open},
but lack depth conditioning and break under large viewpoint changes.
Our method, on the other hand, explicitly models cross-view consistency by optimizing \ourmethod objective and generalizes to arbitrary viewpoints without imposing layout constraints.

\para{Diffusion Synchronization.}
Several works have explored synchronization techniques for diffusion models~\cite{bartal2023multidiffusionfusingdiffusionpaths,lee2023syncdiffusioncoherentmontagesynchronized,geng2024visualanagramsgeneratingmultiview,liu2023textguidedtexturingsynchronizedmultiview,kim2024synctweediesgeneralgenerativeframework} to achieve consistency across multiple outputs. However, existing approaches primarily address specialized cases: panoramic view generation~\cite{bartal2023multidiffusionfusingdiffusionpaths, lee2023syncdiffusioncoherentmontagesynchronized, kim2024synctweediesgeneralgenerativeframework}, orthogonal transformations~\cite{geng2024visualanagramsgeneratingmultiview}, or object-level consistency with evenly distributed camera positions and well established UV map~\cite{kim2024synctweediesgeneralgenerativeframework, liu2023textguidedtexturingsynchronizedmultiview}. Our work addresses a fundamentally different challenge: maintaining consistent image generation across scene-level camera trajectories with arbitrary viewpoints and significant perspective variations.
% In this complex setting, naive averaging of denoising features fails to preserve cross-view consistency due to the non-orthogonal nature of the transformations and the extensive variance in viewpoint geometry.

\para{Diffusion Inference-Time Optimization.}
To circumvent the need for large-scale model fine-tuning, several approaches have explored test-time optimization with directional guidance. Diffusion-TTA~\cite{prabhudesai2023Diffusion-TTA} adapts discriminative models using pre-trained generative diffusion models, updating parameters through gradient backpropagation at inference time. Other methods~\cite{novack2024DITTO,tsai2024gdageneralizeddiffusionrobust,phung2023groundedtexttoimagesynthesisattention} compute directional controls for single-instance generation (music, images) by calculating directional losses during diffusion and back-propagating to update denoising features. While these methods focus on intra-level control with easily quantifiable directions, our approach is the first to propose multi-channel test-time optimization for cross-view control using self-provided directional guidance, maintaining consistency across arbitrary viewpoints without requiring additional training.
\begin{figure*}[t]
  \centering \small
   \includegraphics[width=0.94\linewidth]{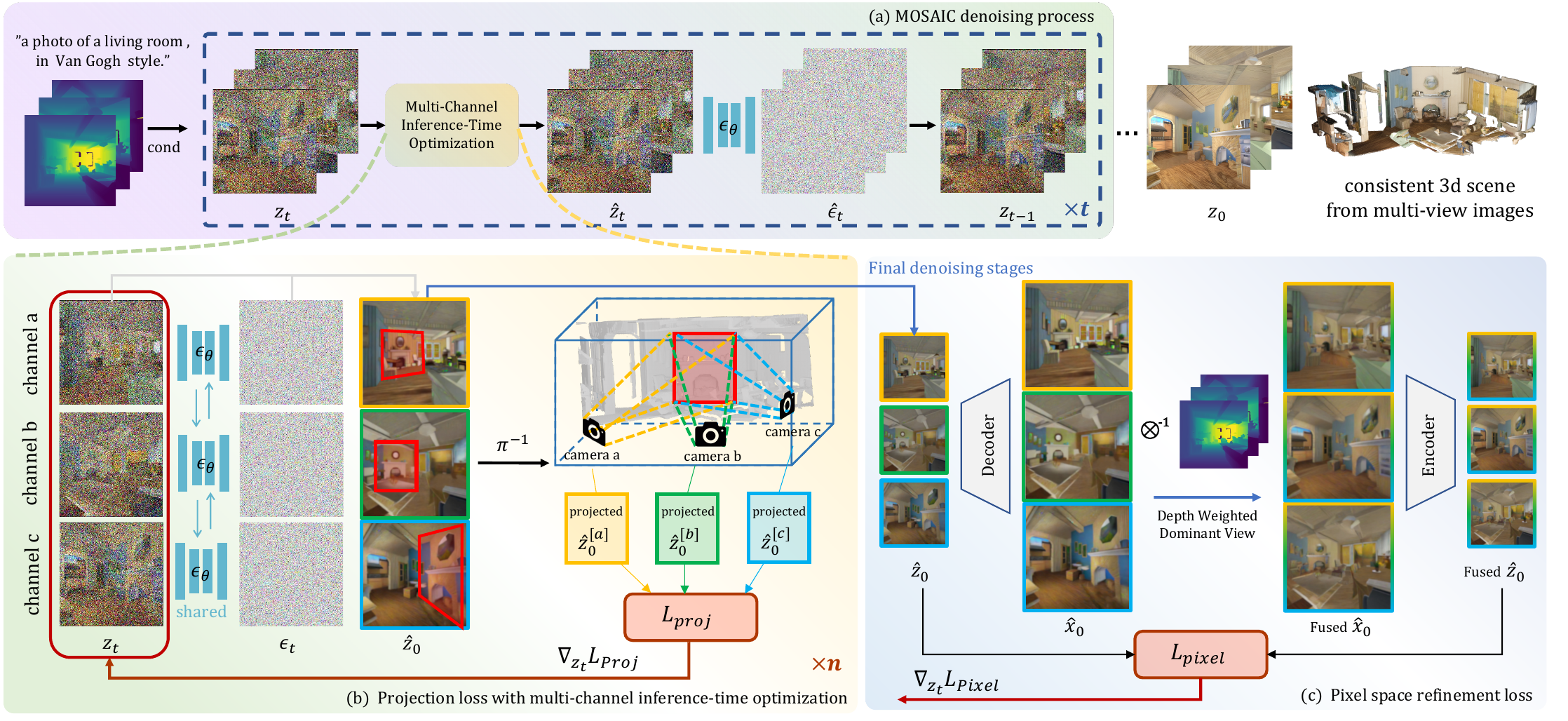}
   \vspace{-0.2cm}%
   \caption{\textbf{MOSAIC overview.} 
(a) \emph{Multi-channel denoising.}  Each depth–text–conditioned view is assigned its own latent channel.  A shared denoiser iteratively refines the latent set while a multi-channel inference-time optimizer keeps the channels synchronized.  
(b) \emph{Projection loss.}  At every step the predicted clean latents $z_{0}$ guided depth-weighted projection loss $L_{\text{proj}}$ drives the channels toward a geometry-consistent solution.  
(c) \emph{Pixel-space refinement.}  During the final denoising stages, the pixel-level loss $L_{\text{pixel}}$ fuses the views and enforces RGB consistency, yielding photorealistic, cross-view-aligned images that can be reconstructed into a coherent 3-D scene.}
   \label{fig:overview}%
   \vspace{-0.5cm}%
\end{figure*}

\begin{figure*}[t]
  \centering
   \includegraphics[width=\linewidth]{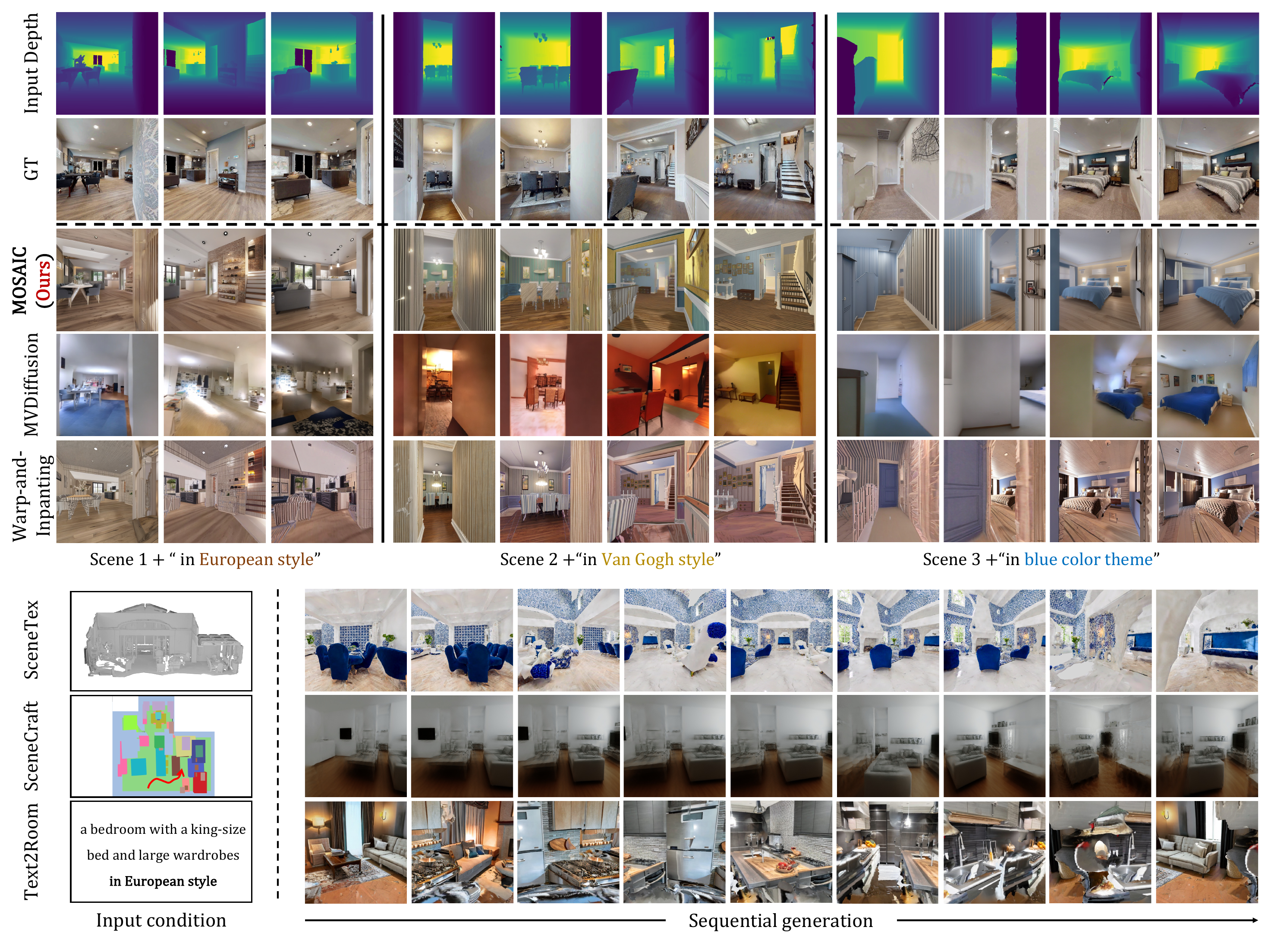}
   \vspace{-0.8cm}
   \caption{\textbf{Qualitative comparison with multi-view baselines.}  
For three indoor scenes (each conditioned on a style prompt) we show: input depth maps, ground-truth RGB, our MOSAIC result, and two baselines (MVDiffusion, Warp-and-Inpainting) sharing the identity input format.  
Below, we compare against baselines (SceneTex, SceneCraft, Text2Room) using their native inputs.  
MOSAIC maintains photorealism, cross-view consistency, and prompt fidelity, whereas competing methods exhibit blur, style drift, or geometric artifacts.}%
   \label{fig:results_baseline}%
   \vspace{-0.5cm}
\end{figure*} 

\section{Problem Definition}
\label{sec:problem_def}
We seek to acquire digital replicas of real-world, multi-room indoor environments in a privacy-preserving manner.
To safeguard privacy, we refrain from collecting sensitive real RGB data and capture only geometric structures instead.
From these structures, we generate \textit{synthetic} RGB data---producing complex, photorealistic scenes that closely align with reality while ensuring no actual environmental details are directly revealed. 
In this work, we assume that the geometric structures are collected as multiple depth images $\{d^{[1]},\dots,d^{[N]}\}$ by mobile robots deployed to real-world scenes.
To fully cover multi-room environments, the mobile robot plans proper camera pose sequences to ensure overlapping views.
Namely, for each $d^{[i]}$,  there must exist some area that is also covered by at least one other view $d^{[j]}$ $(j\neq i)$.
From $\{d^{[i]}\}_{i\in[N]}$, we are interested in generating corresponding RGB images $\{x^{[i]}\}_{i\in[N]}$ that are multi-view consistent: overlapping depth views must lead to consistent RGB outputs where they overlap in the 3D scene ${\bar{x}}$.
Namely, for each pixel $p$ in ${\bar{x}}$ that is covered by multiple depth views, i.e., $I_p := \{i \in [N] \mid  p \in d^{[i]}\}$, the RGB output should agree:
\begin{equation}\label{eq:consistent_rgb_goal}
    x^{[i]}[p] = x^{[j]}[p] = {\bar{x}}[p], \forall i, j\in I_p, i\neq j
\end{equation}
Then the generated RGB would form a complete 3D multi-room scene ${\bar{x}}$ once warped by the cameras poses $\{T^{[i]}\}_{i\in[N]}$ (e.g., ${\bar{x}} = \sum_{i\in[N]} T^{[i]} x^{[i]}$).

\section{Method: MOSAIC}
\label{sec:method}
\subsection{Preliminaries}
Denoising Diffusion Implicit Models (DDIM) \cite{ddim} train a generative model $p_\theta(z_0)$ to approximate a data distribution $q(z_0)$ given samples $z\in\mathcal{Z}$ from $q$.
DDIM considers the following non-Markovian inference model:
\begin{equation}\label{eq:ddim_q}
    q_\sigma\left(z_{1: T} \mid z_0\right):=q_\sigma\left(z_T \mid z_0\right) \prod_{t=2}^T q_\sigma\left(z_{t-1} \mid z_t, z_0\right)
\end{equation}
where $\sigma\in\mathbb{R}_{\geq 0}^T$ is a real vector and $z_{1:T}$ are latent variables in $\mathcal{Z}$.
The inference procedure $q_\sigma\left(z_{t-1} \mid z_t, z_0\right)$ is parameterized by a decreasing sequence $\alpha_{1:T}\in(0,1]^\top$.
To approximate $q(z_0)$, DDIM learns a generative process
\begin{equation}\label{eq:ddim_p}
     \quad p_\theta\left(z_{0: T}\right):=p_\theta\left(z_T\right) \prod_{t=1}^T p_\theta^{(t)}\left(z_{t-1} \mid z_t\right).
\end{equation}
Starting from a prior $p_\theta(z_T)=\mathcal{N}(0, I)$, $z_t$ is sampled by $p_\theta^{(t)}(z_{t-1}\mid z_t)=q_\sigma\left(z_{t-1} \mid z_t, f_\theta^{(t)}(z_t)\right)$, where 
\begin{equation}
    f_\theta^{(t)}\left(z_t\right):=\left(z_t-\sqrt{1-\alpha_t} \cdot \epsilon_\theta^{(t)}\left(z_t\right)\right) / \sqrt{\alpha_t}
\end{equation}
predicts $z_0$ with a noise prediction model $\epsilon_\theta^{(t)}$.
% $p_\theta^{(t)}(z_{0}\mid z_1)$ is then $\mathcal{N}(f_\theta^{(1)}(z_1),\sigma_1^2 I)$.
Learning is performed by optimizing the standard variational objective
\begin{equation}\label{eq:ddim_elbo}
    J_\sigma\left(\epsilon_\theta\right):=\mathbb{E}_{z_{0: T} \sim q_\sigma}\left[\log q_\sigma\left(z_{1: T} \mid z_0\right)-\log p_\theta\left(z_{0: T}\right)\right]
\end{equation}

In this work, we generate RGB images from depth views $d$, via ControlNet \cite{control_net}: $\epsilon_\theta^{(t)}(z_t,d)$.
In addition, we sample latent RGB representations $z$ from $p_\theta$ which can later be decoded into full images via a pre-trained VAE decoder~\cite{kingma2022autoencodingvariationalbayes}.

In short, the generation process we consider is
\begin{align}\label{eq:independent_sampling}
    z_0^{[i]}\sim p_\theta(z_0, d^{[i]}), ~ x^{[i]} = g(z_0^{[i]}) ~ \text{for} ~ i \in [N].
\end{align}
For clarity, we drop the depth dependency $d^{[i]}$ in notations in the rest of this paper. While the generation of a single image $x_i$ can be readily solved via depth-conditioned diffusion models~\cite{control_net}, the generation of complete scenes from multiple depth images remains unsolved.

\subsection{MOSAIC Formulation}
Intuitively, the sampling procedure in \cref{eq:independent_sampling} would generate inconsistent RGB views because the generative process $p_\theta$ is trained with independent samples $z_0\sim q(z_0)$.
In our case, however, the samples $z_0^{[1:N]}$ are indeed \textit{dependent} since they are taken from the same complete scene ${\bar{z}_0}$.
To fundamentally address the consistency issue, our key insight is to \textit{explicitly model such dependency by incorporating extra projection-based conditionals} $q(z_0^{[i]}\mid {\bar{z}_0})$ in the inference procedures.
In practice, given the scene in the latent representation ${\bar{z}_0}$, $z_0^{[i]}$ can be collected by sampling a camera pose with projection function $\pi^{[i]}$ and setting $z_0^{[i]} = \pi^{[i]}({\bar{z}_0})$; subsequent latents $z_{1:T}^{[i]}$ can be sampled as usual.
% See \cref{fig:compare_single_multi_branch} for an illustraion.
The updated inference procedure for $N$ depth views from the same scene ${\bar{z}_0}$ can be rewritten from~\cref{eq:ddim_q} as:
\begin{equation}\label{eq:mosaic_q}
    q_\sigma(z_{0:T}^{[1:N]}\mid {\bar{z}_0}) = \prod_{i\in[N]} q(z_0^{[i]}\mid {\bar{z}_0}) q_\sigma (z_{1:T}^{[i]} \mid z_0^{[i]}).
\end{equation}
The corresponding generative process can be extended to
\begin{equation}\label{eq:mosaic_p}
    p_{\theta,\phi}(z_{0:T}^{[1:N]}, {\bar{z}_0}) = p_\phi({\bar{z}_0} \mid z_0^{[1:N]}) \prod_{i\in[N]} p_\theta(z_{0:T}^{[i]}),
\end{equation}
where $\phi$ parameterizes the reverse process of image projection $q(z_0^{[i]}\mid {\bar{z}_0})$.
We name the model $p_{\theta,\phi}$ as \textbf{M}ulti-view \textbf{O}verlapped \textbf{S}cene \textbf{A}lignment with \textbf{I}mplicit \textbf{C}onsistency (\ourmethod), which can be learned by minimizing:
\begin{align}
    & J_{\sigma,\mathrm{\ourmethod}}\left(\epsilon_\theta, \phi\right):= \nonumber\\
    & \mathbb{E}_{(z_{0: T}^{[1:N]},{\bar{z}_0}) \sim q_\sigma}\left[\log q_\sigma\left(z_{0:T}^{[1:N]} \mid {\bar{z}_0}\right)-\log p_{\theta,\phi}\left(z_{0:T}^{[1,N]},{\bar{z}_0}\right)\right]. \label{eq:mosaic_elbo}
\end{align}
% \subsection{Sampling from \ourmethod via Inference-time Optimization}
\subsection{Sampling from \ourmethod with Multi-Channel Inference-Time Optimization}

While \ourmethod captures the dependency between depth views from the same scene, training directly using \cref{eq:mosaic_elbo} is infeasible for two reasons:
(a) optimizing the expectation in \cref{eq:mosaic_elbo} via stochastic batches can be computationally intractable, since sufficient combinations of $N$ views are required even from a single scene. Meanwhile, $N$ should be sufficiently large to effectively cover multi-room environments; and
(b) $p_{\theta,\phi}$ trained with some fixed $N$ cannot easily generalize to various depth view numbers which further limits the applicability.
Hence, we desire a tractable strategy to sample from \ourmethod that is also compatible with an arbitrary number of depth views $N$. Factorizing $q_\sigma$ using \cref{eq:mosaic_q} and $p_{\theta,\phi}$ using \cref{eq:mosaic_p}, we obtain:
\begin{align}
    & J_{\sigma,\mathrm{\ourmethod}}\left(\epsilon_\theta, \phi\right) \nonumber \\
    & \equiv \textstyle\sum_{i\in[N]}  \underbrace{\mathbb{E}_{q_\sigma} \Big[ \log q_\sigma \left(z_{1:T}^{[i]} \mid z_0^{[i]}\right) - \log p_\theta\left(z_{0:T}^{[i]}\right)  \Big]}_{J_\sigma^{[i]}(\epsilon_\theta) \cref{eq:ddim_elbo}} \nonumber \\
    & ~~~~~ - \mathbb{E}_{q_\sigma}\big[ \log p_\phi({\bar{z}_0} \mid z_0^{[1:N]}) \big].\label{eq:mosaic_elbo_factorized}
\end{align}
Note that $p_\phi({\bar{z}_0} \mid z_0^{[1:N]})$ essentially approximates the posterior of the projection operation in forward process and evaluates how well the original scene ${\bar{z}_0}$ is reconstructed from separate views with some parameterization $\phi$.
Hence, we can reasonably set $p_\phi$ to have an inverse exponential dependency on the total re-projection error:
\begin{align}
     p_\phi({\bar{z}_0} \mid z_0^{[1:N]}) \propto \exp(-\textstyle\sum_{i}\|{\bar{z}_0} - f_\phi^{[i]}(z_0^{[i]})\|)
\end{align}
where $f_\phi^{[i]}$ projects individual views back to the original and error is considered within the back projected region.
Hence the second term in \cref{eq:mosaic_elbo_factorized} can be written as $\mathbb{E}_{q_\sigma}\big[\textstyle\sum_{i\in[N]}\|{\bar{z}_0} - f_\phi^{[i]}(z_0{[i]})\|\big]$.
Since ${\bar{z}_0}$ is in general unavailable, we can instead minimize the total projected cross-view error:
\begin{align}\label{eq:mosaic_proj_loss}
    J_\mathrm{Proj}(\phi) = \mathbb{E}_{q_\sigma}\big[\textstyle\sum_{i,j}\|f_\phi^{[i]}(z_0^{[i]}) - f_\phi^{[j]}(z_0^{[j]})\|\big] \geq 0
\end{align}
Hence, \cref{eq:mosaic_elbo_factorized} is equivalent to
\begin{align}\label{mosaic_two_term_loss}
    J_{\sigma,\mathrm{\ourmethod}}\left(\epsilon_\theta, \phi\right) \equiv \textstyle\sum_{i\in[N]} J_\sigma^{[i]}(\epsilon_\theta) + J_\mathrm{Proj}(\phi).
\end{align}
% Without real ${\bar{z}_0}$ samples, $p_{\theta,\phi}$ trained with \cref{mosaic_two_term_loss} will generate consistent, high-quality 3D RGB scenes which, at same time, do not reveal real-world entities.

Although it is impractical to directly train $p_{\theta,\phi}$ using \cref{mosaic_two_term_loss}, it is possible to approximate the samples from $p_{\theta,\phi}$ by fine-tuning the output of pre-trained models $p_\theta$, e.g., Latent Diffusion Model (LDM) \cite{rombach2022high}.
% which sufficiently solves \cref{eq:ddim_elbo} in scale.
% To proceed, we analyze how different learning objectives lead to the gap between samples from $p_\theta$ (i.e., pre-trained models) and from $p_{\theta, \phi}$ (i.e., ideal \ourmethod model).
% To achieve that, we first observe a similar inference chain\footnote{We slightly generalize the notion of ``chain'' outside Markovian cases.} $q_\sigma(z_{1:T}\mid z_0)$ used to generate sample for \ourmethod in \cref{mosaic_two_term_loss} and LDM in \cref{eq:ddim_elbo}.
We observe that for \ourmethod, $p_{\theta,\phi}$ fits trajectories $z_{1:T}^{[1]},\dots,z_{1:T}^{[N]}$  that are mutually dependent at each step $t=1,\dots,T$ through $z_0^{1:N}$, while such dependency is missing from LDM $p_\theta$.
Furthermore, this dependency is exactly captured by $J_\mathrm{Proj}(\phi)$.
If $p_{\theta, \phi}^*$ optimally solves $J_{\sigma,\mathrm{\ourmethod}}$ and $z_0^{[1:N]}\sim p_{\theta, \phi}^*$ , there must exist some ${\bar{z}_0}$ and corresponding projections $\pi^{[1:N]}$ such that $z_0^{[i]}=\pi^{[i]}({\bar{z}_0}),~\forall i\in[N]$.
Then, $J_\mathrm{Proj}(\phi)$ must have been minimized to $0$ with $f_{\phi^*}^{[i]} = (\pi^{[i]})^{-1},~\forall i$.
In other words, samples from an ideal \ourmethod model should have zero projected cross-view error in expectation as long as $f_\phi^{[i]}$ exactly matches the inverse projection $(\pi^{[i]})^{-1}$.
While it is generally hard to fit an inverse projection, it is unnecessary in our case because the ground truth projections $\pi^{[1:N]}$ are directly available when collecting depth conditions $d^{[1:N]}$\footnote{With mobile robots, $\pi^{[i]}$ can be computed from the camera/robot poses in the world coordinate when taking each depth view $d^{[i]}$.}.

Hence, we can approximately sample from \ourmethod by fine-tuning LDM (i.e., $p_\theta$) output at each denoising step $t-1$ by solving the following \textit{inference-time optimization} through gradient decent:
\begin{align}\label{eq:mosaic_optimization}
    \underset{z_{t-1}^{[1:N]}}{\mathbf{min}} & ~ L_\mathrm{Proj}(z_{t-1}^{[1:N]}) ~ \textbf{s.t.} ~ z_{t-1,\text{init}}^{[i]} \sim p_\theta^{(t)}(z_{t-1}^{[i]} \mid z_t^{[i]}),
\end{align}
where the empirical projection loss is given by
\begin{equation}\label{eq:mosaic_optimization_objective}
L_\mathrm{Proj}(z_t^{[1:N]}) = \sum_{i,j}\|(\pi^{[i]})^{-1}(\hat{z}_0^{[i]}) - (\pi^{[j]})^{-1}(\hat{z}_0^{[j]})\|.
\end{equation}
$\hat{z}_0^{[i]}$ is the estimated $z_0^{[i]}$ from $z_t^{[i]}$, i.e., $\hat{z}_0^{[i]} = f_\theta^{(t)}(z_t^{[i]})$.
 
\para{Depth Weighted Projection Loss.} 
When calculating $L_\mathrm{Proj}$, we observe that depth information provides a natural weighting mechanism for view contributions. For points visible from multiple views, the view with smaller depth value (i.e., closest to the camera) likely provides more accurate RGB information due to higher sampling density and reduced occlusion.
To incorporate this insight, we modify our projection loss to weight view contributions based on their relative depth values:
\begin{equation}\label{eq:mosaic_weighted}
L_{\text{Proj}}^{\text{depth}}(z_t^{[1:N]}) = \sum_{i,j} w_{i,j} \cdot \| (\pi^{[i]})^{-1}(\hat{z}_0^{[i]}) - (\pi^{[j]})^{-1}(\hat{z}_0^{[j]}) \|
\end{equation}
\noindent where $w_{i,j}$ weights the importance of consistency between views $i$ and $j$ based on their relative depth values, weights are calculated per pixel. To prioritize views with smaller depth values, we employ weighting scheme:
\begin{equation}
w_{i,j} = \frac{\exp(-\alpha \cdot(\pi^{[i]})^{-1}d^{[i]})}{\exp(-\alpha \cdot (\pi^{[i]})^{-1}d^{[i]}) + \exp(-\alpha \cdot (\pi^{[j]})^{-1}d^{[j]})},
\end{equation}
\noindent where $\alpha$ is a hyperparameter controlling the selectivity of the weighting. With this formulation, when aggregating across all views, the final RGB values for overlapping regions will naturally favor views with minimal depth values.
\subsection{Pixel Space Refinement Loss}
While our projection loss effectively ensures consistency in latent space, the non-linear nature of the VAE~\cite{kingma2022autoencodingvariationalbayes} decoder means that consistency in latent space does not necessarily translate to pixel-space consistency. This discrepancy is particularly pronounced when projection transformations are far from orthogonal. We address this issue through a pixel space refinement process during the final denoising stages.
For each view $i$, we decode its predicted latent into pixel space: $\hat{x}^{[i]} = g(\hat{z}_0^{[i]})$. It is infeasible to calculate $L_{Proj}$ in pixel space, because back-propagation through multi-channel VAE decoder poses computation difficulties. 
We propose to warp these decoded images between views and compute  representations for each view point by depth-weighted fusion of the overlapped views:
\begin{equation}
w_{ij} = \frac{\exp(-\alpha \cdot\pi_{ij} d^{[j]})}{\sum_{k=1}^{N} \exp(-\alpha \cdot\pi_{ik} d^{[k]})}, ~ {x}^{[i]*} = \sum_{j=1}^{N} w_{ij} \cdot \hat{x}^{[j]}
\end{equation}
where $\alpha$ controls the selectivity of weighting and $\pi_{ij}$ is the image warping from $j$ view to $i$ view.
These optimal pixel-space representations are re-encoded to latent space: ${z}_0^{[i]*} = \beta \cdot f({x}^{[i]*})$, where $f$ is the mapping from pixel to latent. The Pixel Space Refinement Loss is defined as:
\begin{equation}
L_{Pixel} = \sum_{i=1}^{N} \| \hat{z}_0^{[i]} - {z}_0^{[i]*} \|
\end{equation}
This directly addresses the latent-to-pixel mapping inconsistencies by forcing latent representations to align with those derived from optimally blended pixel-space images. We utilize the $L_{Pixel}$ at the final denoising stages.

\subsection{Properties of \ourmethod}
% \todo{More views, less variances, better depth and rgb alignment. formulation.} \ruic{Theory proof} \zhixuan{Zhixuan: Experiment results qualitatively and quantitatively}

\para{Training-free inference time scale-up.} The sampling scheme in \cref{eq:mosaic_optimization} essentially consists of two parts: (a) encouraging multi-view consistency via $L_\mathrm{Proj}$ which is defined for an arbitrary number of views $N$; and (b) progressing the reverse process by invoking LDM independently on each view.
Hence, unlike full training using \cref{eq:mosaic_elbo}, sampling from MOSAIC is agnostic of $N$, allowing our generation process to easily scale and adapt to larger scenes with more views during inference time, with no extra training.

\para{Variance reduction for pre-trained LDMs.} Since LDMs are normally pre-trained in scale, we can roughly view them to have zero prediction error in expectation.
However, as will be shown later, LDMs can generate results with varying quality (e.g., in terms of depth preservation) due to the stochasticity of the denoising process, which can cause errors to accumulate until the whole process fails eventually in the warp-inpainting approach.
To this regard, one important advantage of MOSAIC is its ability to \textit{stabilize the denoising process given more overlapping views}.
This can be seen by analyzing the expected variance of a scene $\bar{z}_0$ given a varying number of views $z_0^{[1:N]}$ that overlap.
Consider a partition $z_0^{[1:N]}=\{z_0^{[I_1]},z_0^{[I_2]}\}$ where $I_1 \cup I_2 = \{1,\dots,N\}$,
\begin{align}\label{eq:var_reduce_proof}
    \Sigma(\bar{z}_0 \mid z_0^{[I_1]}) &= \mathbb{E}[\Sigma(\bar{z}_0 \mid z_0^{[I_1]}, z_0^{[I_2]})] + \underbrace{\Sigma(\mathbb{E}[\bar{z}_0 \mid z_0^{[I_1]}, z_0^{[I_2]}])}_{\Sigma \text{~explained by~} z_0^{[I_2]}} \nonumber\\
    &\succeq \mathbb{E}[\Sigma(\bar{z}_0 \mid z_0^{[I_1]}, z_0^{[I_2]})].
\end{align}
Taking expectation over $z_0^{[I_1]}$, we have
\begin{equation}\label{eq:var_reduce_proof_final}
    \mathbb{E}[\Sigma(\bar{z}_0 \mid z_0^{[1:N]})] \preceq \mathbb{E}[\Sigma(\bar{z}_0 \mid z_0^{[I_1]})].
\end{equation}
Namely, when conditioned on more views $z_0^{[I_2]}$, the variance of the final output $\bar{z}_0$ reduces in expectation by an amount positively related to the overlap between $\bar{z}_0$ and $z_0^{[I_2]}$\footnote{The amount reduced represents the variability of $\bar{z}_0$ due to changes in $z_0^{[I_2]}$, which is higher when $z_0^{[I_2]}$ covers more area in $\bar{z}_0$.}.
The same applies when $z_0^{[1:N]}$ is replaced by estimates from intermediate $z_t^{[1:N]}$, hence the variance reduction happens during the entire denoising process.

\section{Experimental Setup}
\label{sec:experiments}
\para{Autonomous Data Collection.} To construct our test set, we deployed an indoor-navigation robot that autonomously explored diverse indoor environments.  
During each trajectory, the robot captured depth maps with its on-board sensors and logged precise camera poses, producing trajectory-aligned inputs compatible with prior work~\cite{tang2023mvdiffusion,yang2025scenecraft,hollein2023text2room,controlnetinpaint2023}.  
We gathered data from 16HM3D~\cite{ramakrishnan2021habitatmatterport3ddatasethm3d} and 5 MP3D~\cite{chang2017matterport3dlearningrgbddata} scenes, for each scene we collected around 10 independent trajectories.  
For captioning, we rendered each depth map to a grayscale image and queried the vision–language model \texttt{GPT-4o}~\cite{brown2020languagemodelsfewshotlearners} to generate concise, scene-aware descriptions, yielding 2011 \emph{(depth, pose, caption)} triplets in total.

\para{Baselines.} We compare our method with state-of-the-art multi-view generators that leverage geometric priors:  
MVDiffusion~\cite{tang2023mvdiffusion}, Warp-Inpaint~\cite{controlnetinpaint2023}, SceneTex~\cite{chen2023scenetexhighqualitytexturesynthesis}, and SceneCraft~\cite{yang2025scenecraft}.  
Although our depth maps can be converted to meshes required by SceneTex, running SceneTex itself to synthesize multi-view images for a single scene takes more than 20 hours, making a full scenes evaluation impractical.  
Therefore, we run SceneTex on an 8-scene subset and report its results separately for fairness.  
SceneCraft expects semantic layouts with per-object depth; we derive the necessary bounding-box layouts from our dataset.  
We additionally assess semantic fidelity against Text2Room~\cite{hollein2023text2room}, a specialist indoor-scene generator.

\para{Evaluation metrics.} Our task seeks images that (i) are geometrically consistent across overlapping views, (ii) exhibit high perceptual quality and stylistic coherence, and (iii) faithfully reflect the text prompt. Following MVDiffusion~\cite{tang2023mvdiffusion}, we measure geometric consistency with Warped PSNR~\cite{5596999}—the PSNR after warping one view into another using ground-truth geometry—and its normalized variant against GT, Warped Ratio, where 1.0 indicates perfect alignment. Image quality is assessed with Kernel Inception Distance (KID)~\cite{binkowski2018demystifying} and CLIPIQA$^{+}$~\cite{wang2023exploring}; prompt adherence is quantified by CLIPScore~\cite{hessel2021clipscore} and CLIPConsistency~\cite{radford2021learning}.
\begin{figure}
\vspace{-0.1cm}
    \centering
    \includegraphics[width=0.9\columnwidth]{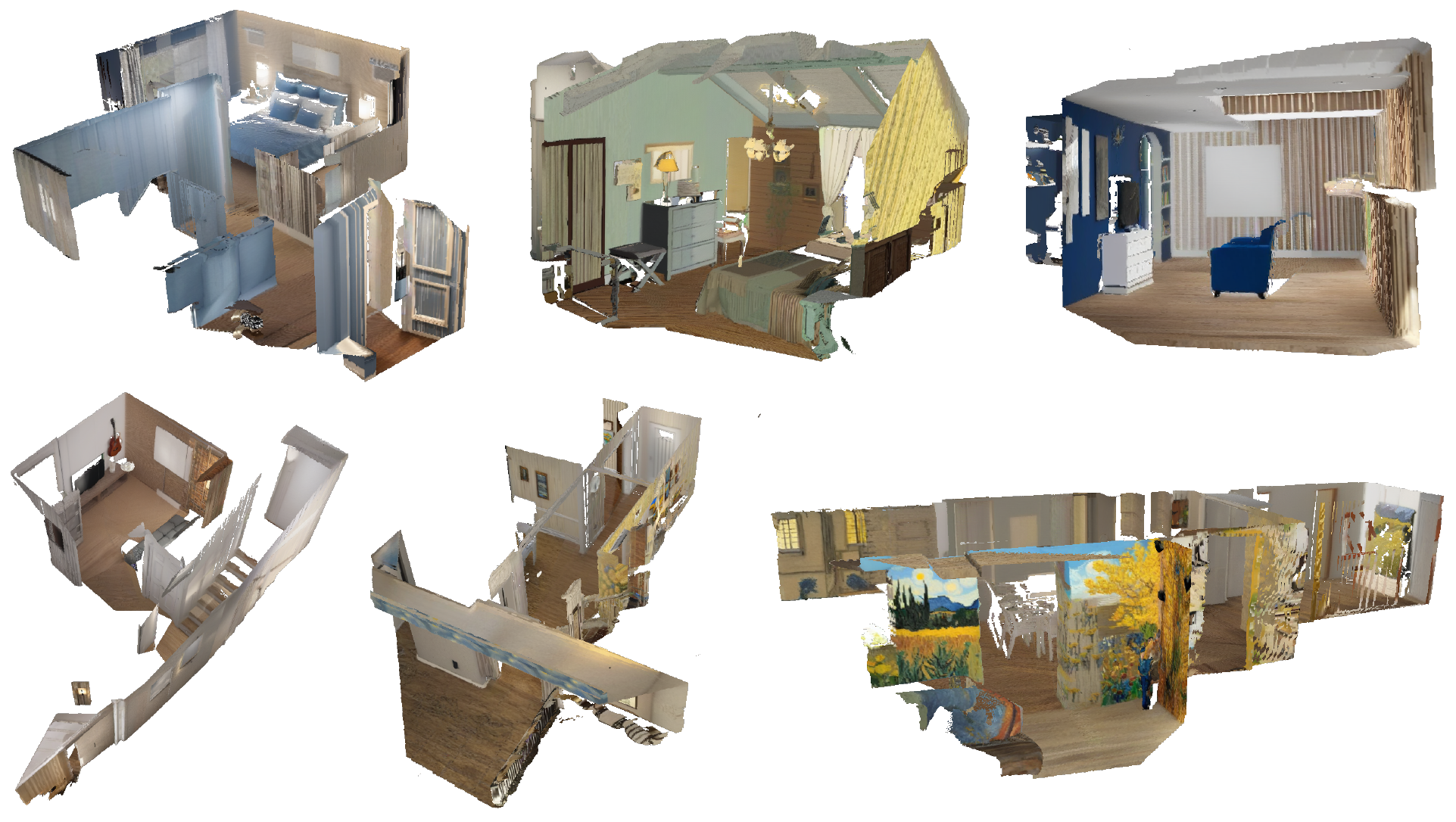}
    \vspace{-0.1cm}
    \caption{\textbf{Scene-level reconstruction.} Fusing the multi-view images generated by MOSAIC with a standard TSDF pipeline produces coherent, textured meshes across diverse indoor rooms.}
    \label{fig:mesh}
    % \vspace{-0.3cm}
\end{figure}
\begin{table}[t]
\footnotesize
\centering
\caption{Quantitative evaluation with Kernel Inception Distance (KID), CLIPIQA$^{+}$, CLIPScore (CS) and CLIPConsistency(CC).}
\label{tab:baseline_image_qual}
% \vspace{-0.2cm}
\resizebox{\columnwidth}{!}{
\begin{tabular}{l|cccc}
% \begin{tabular}{p{2.5cm}|p{1.3cm}p{1.3cm}p{1.3cm}p{1cm}}
\toprule
Method      & KID $\downarrow$ & CS $\uparrow$ & CIQA $\uparrow$ & CC $\uparrow$ \\ \midrule
Warp-Inpaint \cite{controlnetinpaint2023} & 0.04646          & 0.6849          & 0.6517          & 29.93          \\
MVDiffusion \cite{tang2023mvdiffusion}                                                     & 0.03640          & 0.7016          & 0.6025          & 29.41          \\
SceneCraft \cite{yang2025scenecraft}                                                      & 0.07697          & 0.7011          & 0.4531          & 27.68          \\
Text2Room \cite{hollein2023text2room}                                                       & 0.08594          & 0.6618          & 0.4388          & 27.34          \\
\ourmethod (ours)                                                      & \textbf{0.03391} & \textbf{0.7166} & \textbf{0.6526} & \textbf{30.85} \\ 
\midrule
SceneTex \cite{chen2023scenetexhighqualitytexturesynthesis} - small set & 0.10536 & 0.7045 & 0.6308 & 28.5740          \\
\ourmethod - small set (ours)                                                     & \textbf{0.06547} & \textbf{0.7141} & \textbf{0.7322} & \textbf{29.5043} \\ 
\bottomrule
\end{tabular}
}
% \vspace{-15pt}
\end{table}
% \begin{table}[t]
% \footnotesize
% \centering
% \caption{Quantitative evaluation with Kernel Inception Distance (KID), CLIPIQA$^{+}$, CLIPScore (CS) and CLIPConsistency(CC).}
% \label{tab:baseline_image_qual}
% \vspace{-0.2cm}
% \resizebox{\columnwidth}{!}{
% \begin{tabular}{l|cccc}
% % \begin{tabular}{p{2.5cm}|p{1.3cm}p{1.3cm}p{1.3cm}p{1cm}}
% \toprule
% Method      & KID $\downarrow$ & CS $\uparrow$ & CIQA $\uparrow$ & CC $\uparrow$ \\ \midrule
% Warp-Inpaint & 0.04646          & 0.6849          & 0.6517          & 29.93          \\
% MVDiffusion                                                    & 0.03640          & 0.7016          & 0.6025          & 29.41          \\
% SceneCraft                                                & 0.07697          & 0.7011          & 0.4531          & 27.68          \\
% Text2Room                                                  & 0.08594          & 0.6618          & 0.4388          & 27.34          \\
% ours                                                     & \textbf{0.03391} & \textbf{0.7166} & \textbf{0.6526} & \textbf{30.85} \\ 
% \bottomrule
% \end{tabular}
% }
% \vspace{-15pt}
% \end{table}

\section{Results}
\label{sec:results}

\subsection{Evaluation Against Baselines}
\paragraph{Qualitative Comparisons.} As illustrated in Fig.~\ref{fig:results_baseline}, \ourmethod{} generates scenes that are simultaneously more photorealistic and more cross-view–coherent than all competing methods when provided with the same depth inputs.  
MVDiffusion~\cite{tang2023mvdiffusion} exhibits clear view-to-view inconsistencies and is unable to accommodate stylistic prompts such as ``in Van Gogh style.’’  
The warp-and-inpainting pipeline~\cite{controlnetinpaint2023} produces plausible first frames, yet progressive style drift and error accumulation degrade subsequent views.  
SceneCraft~\cite{yang2025scenecraft}, hindered by its NeRF backbone, fails to reconstruct complex layouts and introduces pronounced blur.  
Text2Room~\cite{hollein2023text2room}, which also relies on warp-and-inpainting, shows patchwork seams and depth misalignment.  
SceneTex~\cite{chen2023scenetexhighqualitytexturesynthesis} requires over 20\,hours to process a single scene, whereas \ourmethod finishes in just a few minutes. Moreover, SceneTex’s score distilling optimization often converges to a single appearance mode, yielding nearly identical textures across seeds; our DDIM sampler preserves diversity under identical prompts.  
Overall, MOSAIC preserves depth fidelity and stylistic intent across viewpoints and eliminates boundary artifacts through its depth-weighted projection mechanism.

\para{Quantitative Comparisons.} 
Tab.~\ref{tab:baseline_image_qual} quantitatively compares our approach with Warp-and-Inpainting~\cite{controlnetinpaint2023}, MVDiffusion~\cite{tang2023mvdiffusion}, SceneCraft~\cite{yang2025scenecraft}, Text2Room~\cite{hollein2023text2room}, and SceneTex~\cite{chen2023scenetexhighqualitytexturesynthesis}. \ourmethod achieves the lowest KID ($\downarrow$\,0.0391) and the highest CS ($\uparrow$\,0.7166), CIQA (0.6526), and CC (30.85), outperforming every baseline across all perceptual metrics; it also surpasses SceneTex on the 8-scene subset.  
Tab.~\ref{tab:baseline_psnr} further highlights our superior geometric consistency, recording a Warped PSNR of 25.30 and a Warped Ratio of 0.99—virtually matching ground truth (25.45, 1.00) and eclipsing all alternatives.  
Although MVDiffusion attains a competitive KID of 0.0364, its Warped PSNR is only 13.58, underscoring its limited multi-view consistency in 3D.  
In short, MOSAIC delivers state-of-the-art performance in both image fidelity and cross-view alignment.
\begin{figure}
% \vspace{-0.5cm}
    \centering
    \includegraphics[width=\columnwidth]{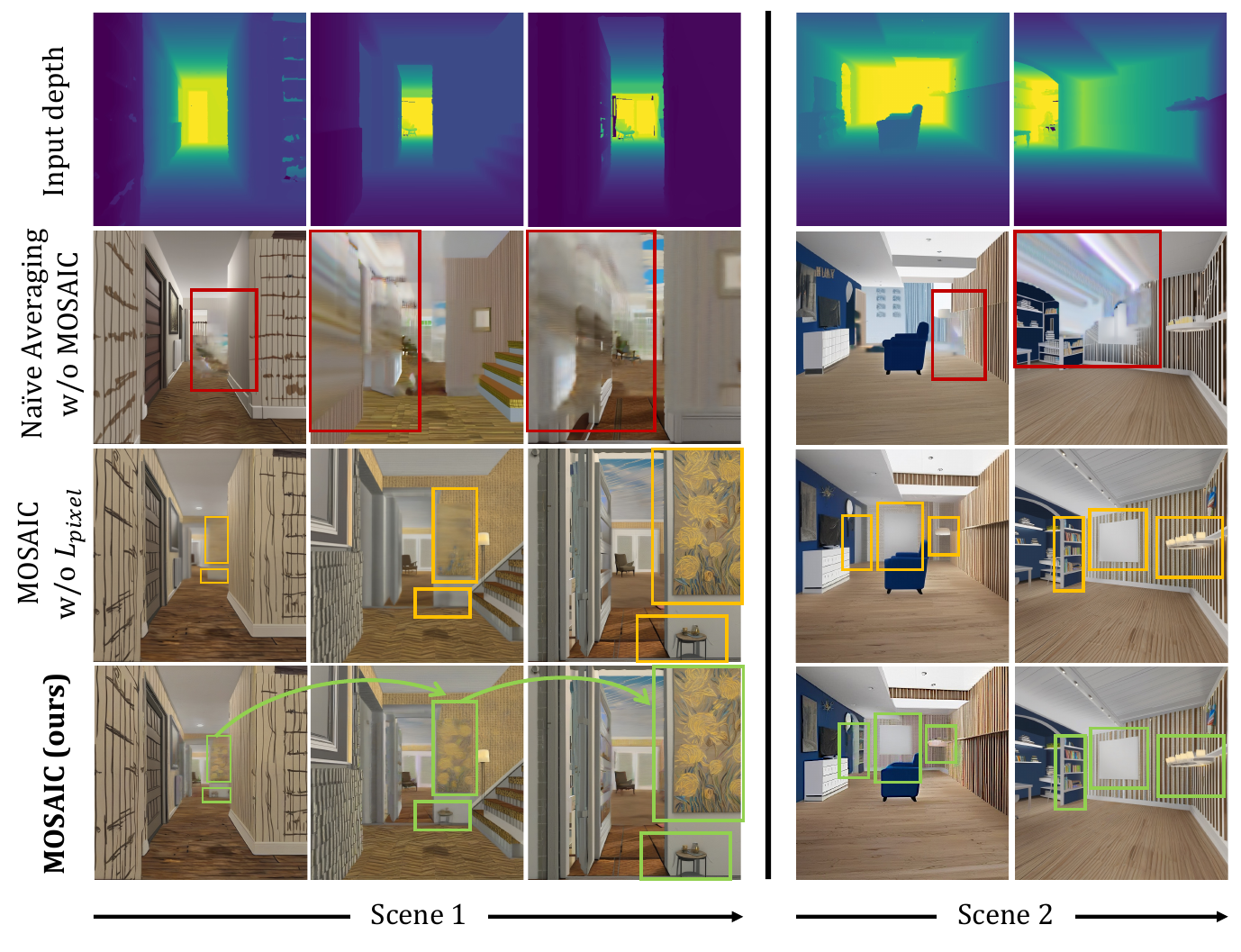}
    % \vspace{-0.8cm}
    \caption{\textbf{Qualitative ablation.}  
Columns show matched views for two scenes; rows compare naïve averaging, MOSAIC without $L_{\text{pixel}}$, and full MOSAIC. Boxes indicate identity objects across different viewpoints along generation.
\textcolor{red}{Red}: blur/ghosts artifacts; \textcolor{orange}{orange}: inconsistent texture drift; \textcolor{green}{green}: full model corrects both.}
    \label{fig:ablation}
    \vspace{-0.3cm}
\end{figure}
% \begin{table}[t]
% \small
% \centering
% % \vspace{-0.3cm}%
% \caption{Cross-view consistency analysis.}
% \label{tab:baseline_psnr}
% \vspace{-0.2cm}
% \resizebox{0.9\columnwidth}{!}{
% % \begin{tabular}{p{2.5cm}|p{3cm}p{2.5cm}}
% \begin{tabular}{l|cc}
% \toprule
% Method            & Warped PSNR ↑  & Warped Ratio ↑ \\ \midrule
% GT          & 25.45          & 1.00           \\ \midrule
% MVDiffusion~\cite{tang2023mvdiffusion} & 13.58          & 0.53           \\
% Warp-Inpaint~\cite{controlnetinpaint2023}  & 22.00          & 0.86           \\
% Ours Pixel  & \textbf{25.30} & \textbf{0.99}  \\ \bottomrule
% \end{tabular}
% }
% \vspace{-0.5cm}%
% \end{table}

\begin{table}[t]
\footnotesize
\centering
\caption{Cross-view geometric consistency analysis.}
\label{tab:baseline_psnr}
\renewcommand{\arraystretch}{0.9}
\begin{tabular}{l|cc}
\toprule
Method            & Warped PSNR ↑  & Warped Ratio ↑ \\ 
\midrule
GT          & 25.45          & 1.00           \\ 
\midrule
MVDiffusion~\cite{tang2023mvdiffusion} & 13.58          & 0.53           \\
Warp-Inpaint~\cite{controlnetinpaint2023}  & 22.00          & 0.86           \\
MOSAIC (ours)  & \textbf{25.30} & \textbf{0.99}  \\ 
\bottomrule
\end{tabular}
\vspace{-0.5cm}
\end{table}

\vspace{-0.1cm}
\subsection{Ablation Study}

% \begin{figure}
% \vspace{-0.3cm}
%     \centering
%     \includegraphics[width=\columnwidth]{img/ablation2.pdf}
%     \vspace{-0.4cm}
%     \caption{Qualitative ablation results. \textcolor{red}{Red}: artifacts; \textcolor{Orange}{orange}: pixel space texture misaligned; \textcolor{green}{Green}: with pixel space loss, full model (\ourmethod) aligns the pixel space color.}
%     \label{fig:ablation}
%     \vspace{-0.3cm}
% \end{figure}
\para{Effect of MOSAIC Objective and Multi-Channel Test-Time Optimization.}
We build a naive ablation by extending~\cite{kim2024synctweediesgeneralgenerativeframework} to scene-level multi-view image generation, averaging each denoising latent prediction $z_0$ without our multi-channel inference time optimization for \ourmethod objectives. As reported in Tab.~\ref{tab:ablation}, this variant records the worst KID and the weakest consistency scores. 
Fig.~\ref{fig:ablation}\,(second row) further exposes severe blur and cross-view misalignment (red boxes).  
Conversely, the full MOSAIC model achieves the best KID, the strongest consistency, and visually tight alignment across views (green boxes), confirming that \ourmethod is critical for large viewpoint changes naive averaging collapses both appearance and geometry.
% confirming that our objective is critical for multi-room scenes with large viewpoint changes where naive averaging collapses both appearance and geometry.

\para{Effect of Pixel Space Refinement.}
Adding the pixel-space loss $L_{pixel}$ markedly improves performance (Tab.~\ref{tab:ablation}), lowering KID and pushing consistency scores close to ground truth.  
In Fig.~\ref{fig:ablation}, boxes track identical objects across different viewpoints: without $L_{pixel}$ (third row) the flower vanishes, the stool disappears, and decorative shapes deform (orange boxes).  
With $L_{pixel}$ (fourth row) textures remain coherent from different viewpoints (green boxes), demonstrating the benefit of pixel-level supervision.

\para{Effect of Trajectory Key Frame Selection.} 
Tab.~\ref{tab:ablation} contrasts our region-coverage sampling with uniform trajectory sampling.  
Although overall image quality and consistency are similar, text–image alignment degrades under uniform sampling.  
Uniformly selected frames often face blank walls, leaving the VLM with insufficient semantic cues; it is also likely to cause pretrained ControlNet~\cite{control_net} branches to hallucinate textures.  
Our key-frame strategy strikes a balance between information-rich viewpoints and sufficient overlap, yielding more accurate text-conditioned results.

\begin{figure}
\vspace{-0.3cm}
    \centering
    \includegraphics[width=\columnwidth]{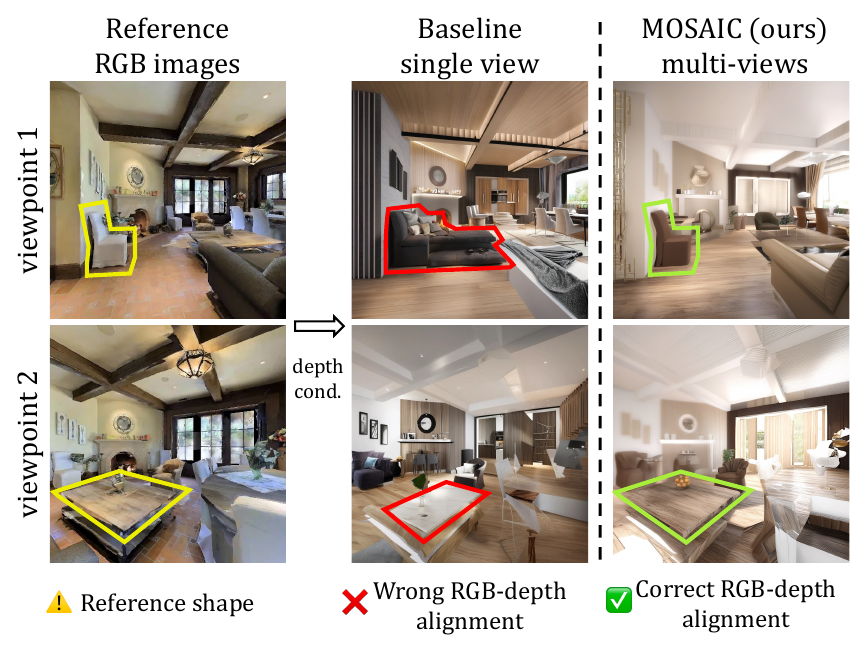}
    \vspace{-0.5cm}
    \caption{\textbf{RGB–depth alignment} Left: reference images at two viewpoints (\textcolor{yellow}{yellow}).  
Middle: single-view baseline producing geometric artifacts (\textcolor{red}{red}).  
Right: MOSAIC conditioned on multiple depth maps recovers accurate geometry (\textcolor{green}{green}).}
    \label{fig:pgm}
    \vspace{-0.5cm}
\end{figure}

\para{Multi-Views for better RGB-Depth Alignment.} Traditional scene-level generation approaches employ warp-and-inpainting strategies that suffer from error accumulation—initial RGB-depth misalignments compound through sequential generation. 
By contrast, Eq.~\ref{eq:var_reduce_proof_final} shows that conditioning on multiple depth views analytically reduces variance, producing geometry closer to ground truth.  
To validate this, we select key frames whose viewpoints overlap by more than 50\% of their pixels and vary the number~$N$ of such views during generation.  
Tab.~\ref{tab:pgm} corroborates this: the normalized MSE between generated depths (estimated with Depth-Anything~\cite{yang2024depthanythingunleashingpower}) and ground-truth depths decreases monotonically as the number of views increases.  
Fig.~\ref{fig:pgm} visualizes the trend: single-view conditioning (red boxes) hallucinates structures that diverge from ground truth (yellow boxes), whereas multi-view conditioning (green boxes) preserves geometric accuracy.  
Our analysis further reveals an inflection point in the quality-viewpoint curve, with text-image alignment steadily improving across viewpoints, while perceptual quality peaks at 2-3 highly overlapping views. This finding suggests an optimal operating point that balances computational efficiency with generation quality.

\begin{table}[t]
\footnotesize
\setlength{\tabcolsep}{3pt}
\centering
\caption{Ablation: Image Quality and 3D Consistency}
% \vspace{-0.2cm}
\label{tab:ablation}
\resizebox{\columnwidth}{!}{
\begin{tabular}{c|cccc|cc}

\toprule
\multirow{2}{*}{} & \multicolumn{4}{c|}{\textbf{Image Quality}}         & \multicolumn{2}{c}{\textbf{Geometry Consistency}} \\ 
                  & KID $\downarrow$ & CS $\uparrow$ & CIQA $\uparrow$ & \multicolumn{1}{c|}{CC $\uparrow$} & PSNR $\uparrow$       & Ratio $\uparrow$      \\ \midrule
w/o test-time optimization & 0.06715          & \textbf{0.7285} & 0.5725          & 31.08          & 15.74          & 0.6185          \\
w/o key frame selection& 0.03920          & 0.6746          & 0.6733          & 29.25          & 24.67          & 0.9692          \\
w/o $L_{Pixel}$              & 0.04273          & 0.7249          & \textbf{0.6797} & \textbf{31.26} & 23.02          & 0.9045          \\
MOSAIC (ours full) & \textbf{0.03391} & 0.7166          & 0.6526          & 30.85          & \textbf{25.30} & \textbf{0.9940} \\
\bottomrule
\end{tabular}
}
% \vspace{-0.5cm}%
\end{table}

% \begin{table}[]
% \footnotesize
% \centering
% \caption{Assessing the effect of number of depth views, NMSE stands for normalized MSE.}
% \vspace{-0.3cm}%
% \label{tab:pgm}
% % \resizebox{0.9\columnwidth}{!}{
% % \begin{tabular}{p{2.5cm}|p{2cm}p{2cm}p{2cm}}
% \begin{tabular}{c|ccc}
% \toprule
%             & NMSE $\downarrow$  & CS $\uparrow$ & KID $\downarrow$\\ \midrule
% single view  &0.018630 & 0.6874 & 0.06123\\ 
% two views & 0.013453 & 0.6883 & \textbf{0.06022}       \\
% three views  & 0.011961 & 0.6986 &  0.06709     \\
% four views  & \textbf{0.011269} & \textbf{0.7100} & 0.07833 \\ \bottomrule
% \end{tabular}
% % }
% \vspace{-0.6cm}%
% \end{table}

\begin{table}[t]
\footnotesize
\centering
\caption{Assessing the effect of number of depth views, NMSE stands for normalized MSE.}
\label{tab:pgm}
\renewcommand{\arraystretch}{0.9}
\begin{tabularx}{0.9\columnwidth}{p{0.3\columnwidth}|XXX}
\toprule
Views  & NMSE $\downarrow$  & CS $\uparrow$ & KID $\downarrow$\\ 
\midrule
single view  &0.018630 & 0.6874 & 0.06123\\ 
two views & 0.013453 & 0.6883 & \textbf{0.06022}\\
three views  & 0.011961 & 0.6986 &  0.06709     \\
four views  & \textbf{0.011269} & \textbf{0.7100} & 0.07833 \\ 
\bottomrule
\end{tabularx}
\vspace{-0.5cm}
\end{table}
% \vspace{-0.4cm}
\section{Conclusion}
% \vspace{-0.2cm}
We proposed \ourmethod, the first training-free multi-view consistent image generation pipeline that operates at scene level. MOSAIC handles arbitrary numbers of views along any trajectory, adapting to extensive camera viewpoint changes. It employs a depth-weighted projection loss and a pixel-space refinement process to maintain visual coherence across complex multi-room environments, making it well-suited for privacy-sensitive applications. We developed a novel multi-channel inference-time optimization procedure that minimizes cross-view projection errors, which we mathematically prove to be equivalent to the ideal learning objective. We showed theoretically and demonstrated experimentally that by intelligently fusing multi-view information, \ourmethod substantially improves alignment between generated RGB images and ground truth depth and significantly outperforms state-of-the-art methods---addressing the error accumulation issues of current autoregressive multi-view image generation pipelines.
\section*{Acknowledgements}
We are grateful to Yanbo Xu, Zhipeng Bao, Yifan Pu, and Zongtai Li for their helpful comments and discussion.
The project was partly supported
by NSF IIS-2112633. 
{
    \small
    \bibliographystyle{ieeenat_fullname}
    \bibliography{main}
}

\end{document}